\pdfoutput=1

\documentclass[11pt]{article}

\usepackage[final]{acl}
\usepackage{amssymb}
\usepackage{times}
\usepackage{latexsym}

\usepackage[T1]{fontenc}

\usepackage[utf8]{inputenc}

\usepackage{microtype}

\usepackage{inconsolata}
\usepackage{algorithm}
\usepackage{algorithmic}
\usepackage{graphicx}
\usepackage{booktabs} 
\usepackage{array} 
\usepackage{colortbl} 
\usepackage{makecell} 
\usepackage{geometry} 
\usepackage{booktabs} 
\usepackage{array} 
\usepackage{colortbl} 
\usepackage{makecell} 
\usepackage{tabularx} 
\usepackage{multirow}
\usepackage{amsmath} 
\title{Hierarchical-Task-Aware Multi-modal Mixture of Incremental LoRA Experts for Embodied Continual Learning}

\author{
 \textbf{Ziqi Jia\textsuperscript{1,2,\protect\footnotemark[1]}},
 \textbf{Anmin Wang\textsuperscript{1,3,\protect\footnotemark[1]}},
 \textbf{Xiaoyang Qu\textsuperscript{1,\protect\footnotemark[2]}},
 \textbf{Xiaowen Yang\textsuperscript{2}},
 \textbf{Jianzong Wang\textsuperscript{1,\protect\footnotemark[2]}}
\\
\\
 \textsuperscript{1}Ping An Technology (Shenzhen) Co., Ltd., Shenzhen, China,
\\
 \textsuperscript{2}Tsinghua Shenzhen International Graduate School, Tsinghua University, Shenzhen, China,
\\
 \textsuperscript{3}Huazhong University of Science and Technology, Wuhan, China,
\\
 \small{
   \textbf{Correspondence:} \href{mailto:jzwang@188.com}{jzwang@188.com}, \href{mailto:quxiaoy@gmail.com}{quxiaoy@gmail.com}}
}

\begin{document}
\maketitle
\footnotetext[1]{Equal Contribution}
\footnotetext[2]{Corresponding Author}
\begin{abstract}
Previous continual learning setups for embodied intelligence focused on executing low-level actions based on human commands, neglecting the ability to learn high-level planning and multi-level knowledge. To address these issues, we propose the Hierarchical Embodied Continual Learning Setups (HEC) that divide the agent's continual learning process into two layers: high-level instructions and low-level actions, and define five embodied continual learning sub-setups. Building on these setups, we introduce the Task-aware \textbf{M}ixture \textbf{o}f \textbf{I}ncremental \textbf{L}oRA \textbf{E}xperts (Task-aware MoILE) method. This approach achieves task recognition by clustering visual-text embeddings and uses both a task-level router and a token-level router to select the appropriate LoRA experts. To effectively address the issue of catastrophic forgetting, we apply Singular Value Decomposition (SVD) to the LoRA parameters obtained from prior tasks, preserving key components while orthogonally training the remaining parts. The experimental results show that our method stands out in reducing the forgetting of old tasks compared to other methods, effectively supporting agents in retaining prior knowledge while continuously learning new tasks.

\end{abstract}

\section{Introduction}

Recently, embodied intelligence has witnessed remarkable progress \cite{DBLP:conf/acl/ShiSYCL24,DBLP:conf/emnlp/FuQGJDZ24,shridhar2020alfred}. Embodied intelligence integrates key capabilities such as perception, cognition, decision-making, and action, aiming to enable agents to perform complex tasks like household chores \cite{DBLP:journals/tmlr/AgrawalPGL24,yang2024octopus}, navigation \cite{singh2023scene,zheng2024towards,DBLP:conf/iclr/ZhangDSZDTSG24}, and object manipulation \cite{DBLP:conf/emnlp/0001VSK24,li2024league++} within physical environments. Unlike the static environment assumptions common in traditional embodied intelligence research, embodied agents are highly likely to encounter novel behavioral patterns or environments after deployment. Therefore, continual learning is crucial for their adaptation to the dynamically changing real world. Kim et al. \cite{kim2024online} proposed a continual learning setup for embodied intelligence based on instruction following. However, with the emergence of large language models (LLMs), embodied agents demonstrate enhanced autonomous decision-making and environmental understanding \cite{DBLP:conf/iclr/SzotSAMMTMHT24,DBLP:conf/icra/MandiJS24,DBLP:conf/corl/HuangXXCLFZTMCS22}, capable of decomposing abstract human instructions into detailed task plans. Yet, previous continual learning setups did not fully address high-level task planning needs or leverage the full power of LLMs.

\begin{figure*}[t!]
    \centering
    \includegraphics[width=\textwidth]{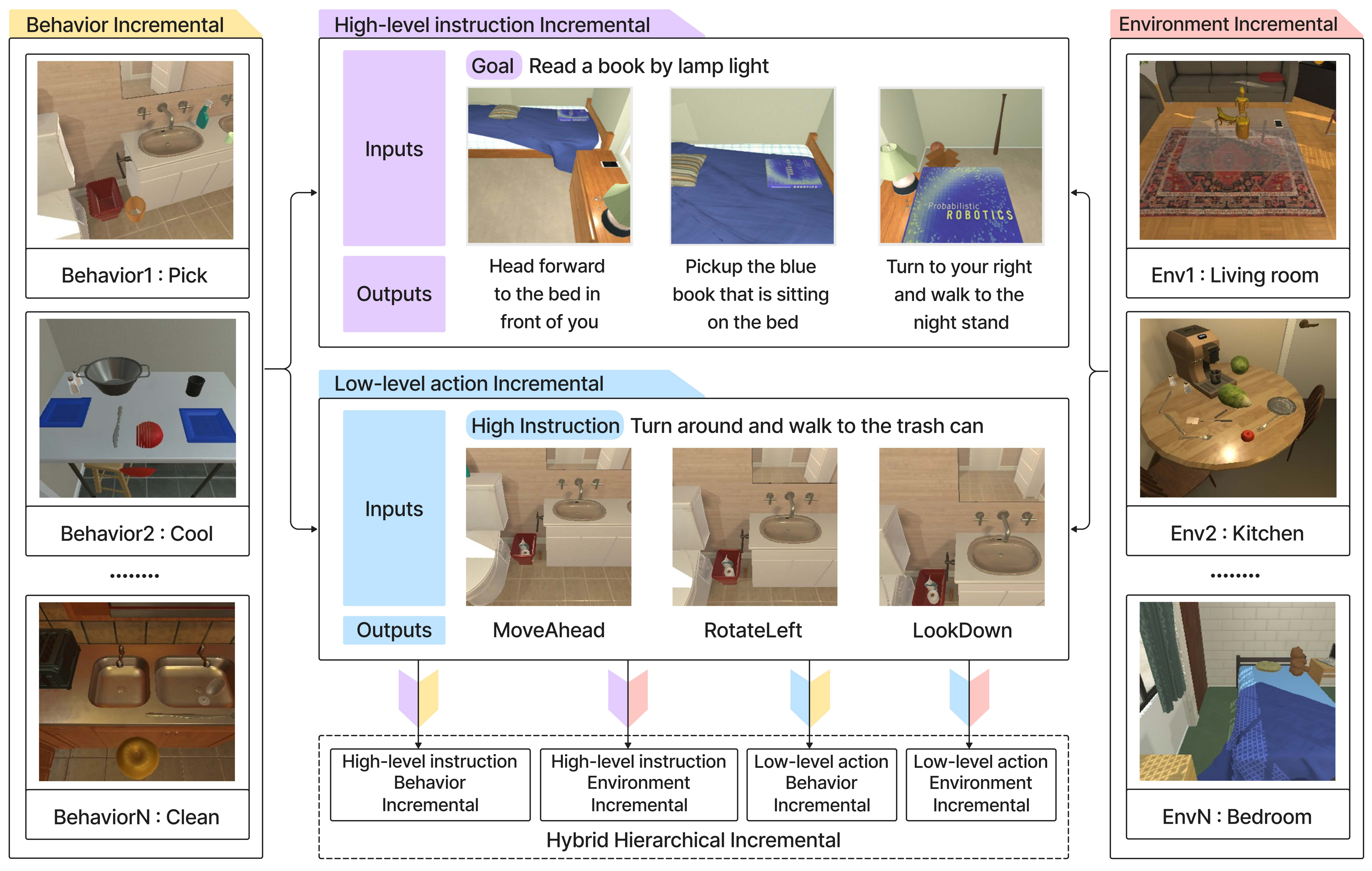}
    \caption{\textbf{Proposed Hierarchical Embodied Continual Learning Setups.} In the High-level Instruction Incremental setup, the agent learns to generate high-level instructions based on goal conditions while preserving prior knowledge. In the Low-level Action Incremental setup, the agent generates specific actions from high-level instructions without forgetting previous knowledge. Combining these with Behavior and Environment Incremental setups, we propose five incremental learning setups. Notably, Hybrid Hierarchical Incremental Learning evaluates the model's ability to retain knowledge across different levels by treating the other four setups as independent tasks for continual learning.}
    \label{fig:1}
\end{figure*}

To address these challenges, we build upon existing embodied intelligence continual learning setups, specifically considering the capabilities of LLMs, further decouples the learning process into two levels: high-level instructions and low-level actions. We propose the Hierarchical Embodied Continual Learning Setups (HEC), which encompasses five specific continual learning setups: High-level Instruction Behavior Incremental Learning, High-level Instruction Environment Incremental Learning, Low-level Action Behavior Incremental Learning, Low-level Action Environment Incremental Learning, and Hybrid Hierarchical Incremental Learning. HEC aims to enable agents to concurrently learn specific actions and high-level task planning abilities, achieving a higher level of autonomy. 


In embodied continual learning, when a model learns a new task, it often suffers from catastrophic forgetting \cite{mccloskey1989catastrophic,goodfellow2013empirical,DBLP:conf/coling/GuF20,ratcliff1990connectionist}, where previously learned knowledge or skills are significantly lost. To address this issue, we propose a Task-aware Mixture of Incremental LoRA Experts (Task-aware MoILE) method. Considering that models in real-world continual learning scenarios typically do not have prior knowledge of task IDs and need to make initial judgments about inputs to assign samples to LoRA experts trained on similar tasks, we first perform clustering analysis on visual-text embeddings processed by a tokenizer and the CLIP model and compute the centroid vector of the cluster to which a new input embedding belongs. This centroid vector represents the fuzzy task category of the current input and is fed into a task-level router to assist the MoE in selecting the appropriate expert. Furthermore, considering the semantic similarities between tasks in embodied continual learning (e.g., "pick" and "place") and to effectively mitigate catastrophic forgetting between different levels (e.g., high-level instructions and low-level actions), we perform singular value decomposition (SVD) on the LoRA parameters trained on previous tasks, retaining the principal components and training the remaining residual components while imposing orthogonality constraints to ensure their orthogonality with the principal components. This approach enables continued training on LoRA experts of similar tasks under the MoE selection mechanism while effectively suppressing catastrophic forgetting and preserving previously acquired skills. Our main contributions include:

\begin{itemize}
    \item We propose the Hierarchical Embodied Continual Learning Setups specifically designed for LLM-driven embodied agents, which decomposes learning into high-level instructions and low-level actions and defines five incremental learning setups.
    \item We introduce a Task-aware Mixture of Incremental LoRA Experts (Task-aware MoILE) continual learning method that effectively mitigates catastrophic forgetting and improves continual learning performance through task-aware expert selection and an SVD-based orthogonal training strategy.
    \item  Evaluations on our HEC demonstrate that our proposed Task-aware MoILE method achieves state-of-the-art results, attaining the highest accuracy and the lowest forgetting measure compared to previous methods.
\end{itemize}
\begin{figure}[ht]
    \centering
   
    \includegraphics[width=0.48\textwidth]{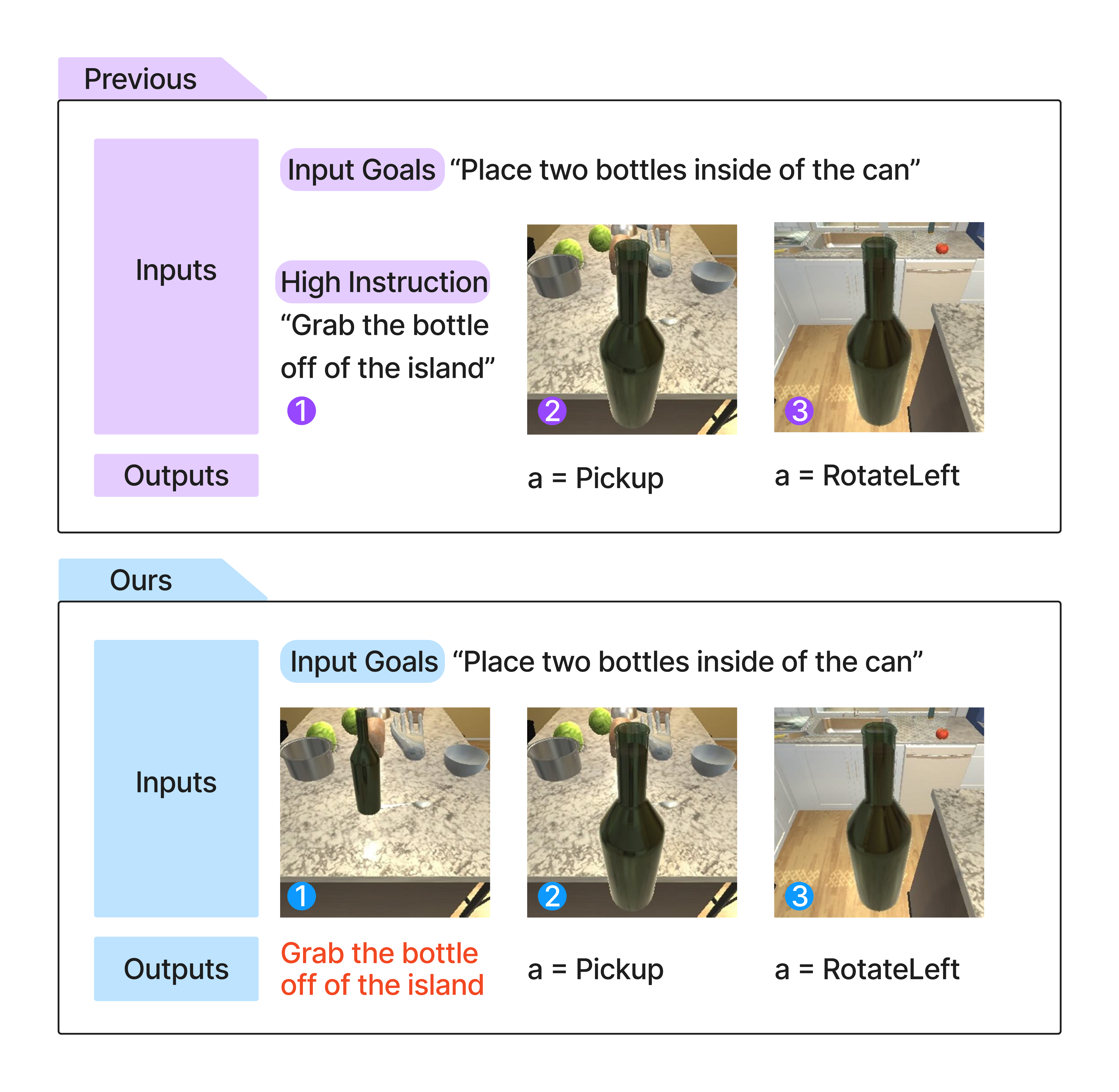}
    \caption{\textbf{Comparison of Our Setup with Previous Work.} Our setup focuses on enabling agents to learn knowledge at different levels, including instruction understanding and action execution, ultimately allowing them to autonomously generate instructions and perform actions in the environment.}
    \label{fig:2}
\end{figure}
\vspace{-1em}
\section{Problem Formulation}
Previous research \cite{kim2024online} has proposed two paradigms for embodied incremental learning: Behavior Incremental Learning and Environment Incremental Learning. These approaches mainly focus on agents following and executing human instructions, while overlooking their autonomous planning and decision-making capabilities. To address this gap, and in light of the growing task planning abilities of Large Language Models and the increasing need for autonomous robotic planning, we propose the Hierarchical Embodied Continual Learning Setups. As illustrated in \textbf{Figure \ref{fig:1}}, HEC consists of five distinct incremental learning modes: High-level Instruction Behavior Incremental Learning (HB), High-level Instruction Environment Incremental Learning (HE), Low-level Action Behavior Incremental Learning (LB), Low-level Action Environment Incremental Learning (LE), and Hybrid Hierarchical Incremental Learning (HH). As illustrated in \textbf{Figure \ref{fig:2}}, HEC emphasizes that during actual task execution, agents not only continually learn execution skills but also continually learn task decomposition and planning.

\subsection{High-level Instruction Incremental Learning}
For high-level instructions, we divide this into two incremental learning modes: High-level Instruction Behavior Incremental Learning, which includes the task set $\mathcal{T}_{\text{HB}}$ = \{EXAMINE, PICK\&PLACE, HEAT, COOL, CLEAN, PICK2\&PLACE, MOVABLE\}; and High-level Instruction Environment Incremental Learning, which includes the environment set $\mathcal{T}_{\text{HE}}$ = \{KITCHENS, LIVINGROOMS, BEDROOMS, BATHROOMS\}. In this setup, we use high-level instructions as labels $y^t_i$ for the $i$-th data point of the $t$-th task, with task goals $\mathcal{G}^t_i$ and scene images $\mathcal{I}^t_i$ as input. Based on the input $x^t_i = \{\mathcal{G}^t_i,\mathcal{I}^t_i\}$ , the model outputs the high-level plan $h^t_i$ for the corresponding task, which is a decomposed sub-task. For example, given the task goal "Read a book under the light," the model will generate different high-level instructions based on different scene images, such as "Pick up the blue book that is sitting on the bed."

\subsection{Low-level Action Incremental Learning}
For low-level actions, we also divide this into two incremental learning modes: Low-level Action Behavior Incremental Learning, which includes the task set $\mathcal{T}_{\text{LB}}$ = \{EXAMINE, PICK\&PLACE, HEAT, COOL, CLEAN, PICK2\&PLACE, MOVABLE\}; and Low-level Action Environment Incremental Learning, which includes the task set $\mathcal{T}_{\text{LE}}$ = \{KITCHENS, LIVINGROOMS, BEDROOMS, BATHROOMS\}. In this setup, we use low-level actions as labels $y^t_i$ for the $i$-th data point of the $t$-th task, with task goals $\mathcal{G}^t_i$, the current high-level instruction $h^t_i$, and scene images $\mathcal{I}^t_i$ as input. Based on the input $x^t_i = \{\mathcal{G}^t_i,h^t_i,\mathcal{I}^t_i\}$, the model outputs the specific action $a^t_i$ for the current task and high-level instruction. For example, given the current high-level instruction "Go to the bookshelf," the model will output specific low-level actions based on the scene image, such as "MoveAhead" and "TurnLeft."

\subsection{Hybrid Hierarchical Incremental Learning}
\begin{figure*}[t!]
    \centering
    \includegraphics[width=\textwidth]{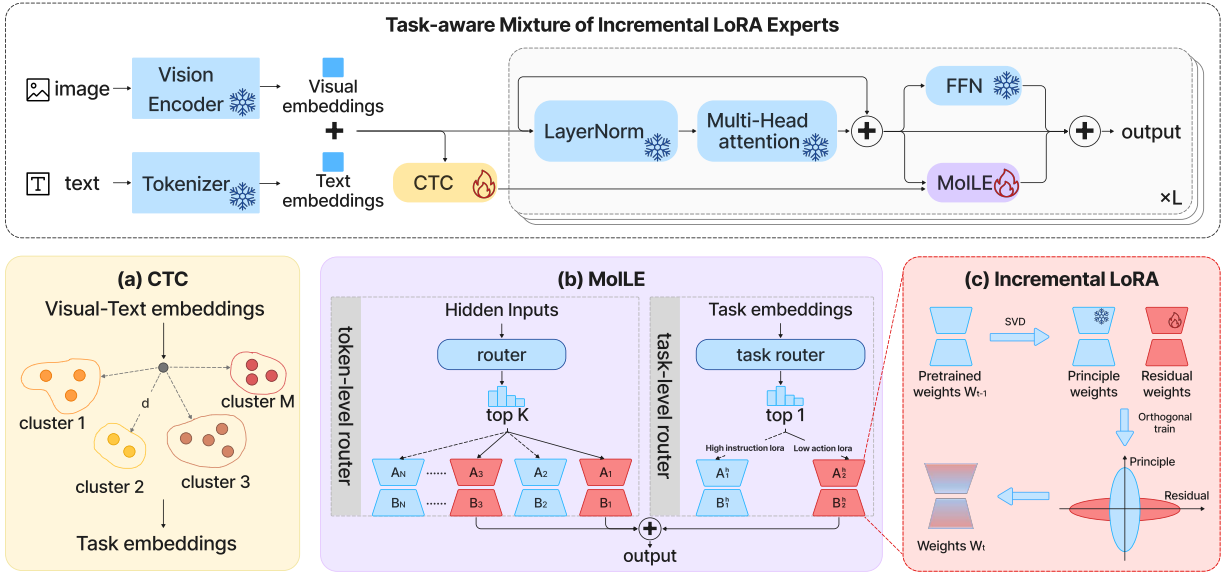}
    \caption{\textbf{The overall architecture of the proposed Task-aware MoILE.} The model transforms Visual-Text embeddings into Task embeddings through (a) CTC, and then performs token-level and task-level expert selection using (b) MoILE. The selected experts are updated via (c) Incremental LoRA, effectively preventing catastrophic forgetting.}
    \label{fig:3}
\end{figure*}
To enable agents to continually learn both high-level instructions and low-level actions in an environment, we propose Hybrid Hierarchical Incremental Learning (HH). We consider the four incremental learning modes mentioned above (LB, HB, LE, and HE) as four independent tasks and learn them sequentially to evaluate the model's forgetting performance when learning tasks at different levels.

\section{Proposed Methods}

To enable the agent to continually learn at both the high-instruction and low-action levels, we propose the method shown in \textbf{Figure \ref{fig:3}}. The method first processes the input embeddings through clustering (3.1), then selects experts at the token-level and task-level using the token-level router and task-level router (3.2). We then perform SVD on the selected LoRA experts, freezing the principal components and training the residuals, ensuring the model retains old knowledge while learning new information (3.3).
\subsection{Cross-modal Task Clustering}

Since directly obtaining task IDs is challenging in real-world scenarios, we use cross-modal task clustering (CTC) to determine the task type of inputs. Specifically, to fully leverage the information from both modalities in the vision-language large model (VLM), we first transform the image and text into feature vectors using pre-trained visual and language encoders, respectively. These feature vectors are then fused or concatenated to generate a unified visual-text embedding vector $x^m$. To allow the cluster centers to dynamically adapt to changes in data distribution, the CTC module updates the positions of the cluster centers with each new batch of samples processed. The specific update formula is as follows:
\begin{equation}
\ c_j^{new} = c_j^{old} + \frac{\alpha}{|S_j^{batch}|} \sum_{x^m_i \in S_j^{batch}} (x^m_i - c_j^{old}),
\end{equation}
where $c_j^{new}$ is the new cluster center, $c_j^{old}$ is the old cluster center, $\alpha$ is the learning rate, $S_j^{batch}$ is the set of samples assigned to cluster center $j$ in the current batch, and $x^m_i$ is the embedding vector of the $i$-th sample.
For each input sample $x^m_i$, calculate its Euclidean distance from all cluster centers $c_l$:
\begin{equation}\label{eq:2}
\ j = \underset {1 \leq l \leq M} { \operatorname {arg\,min} } \|x^m_i - c_l\|^2,
\end{equation}
where $x^m_i$ is the embedding vector of the $i$-th sample, $c_l$ is the $l$-th cluster center, and $M$ represents the number of cluster centers. Once it has been assigned to the nearest cluster center $c_j$, the CTC module outputs the corresponding cluster center vector as the task embedding $e_i$. This task embedding represents the task category of the sample in the multimodal space and is passed to the subsequent task router module.

\subsection{Token-level Router and Task-level Router}
To enable the model to better output based on task hierarchy while maintaining generalization ability in hierarchical embodied continual learning, we introduce two types of LoRA experts in the MoE layer: token-level LoRA and task-level LoRA. Specifically, the token-level router selects the top-$K$ token-level LoRA experts based on the hidden input $x$ from the previous layer. Meanwhile, the task-level router selects the top-1 task-level LoRA expert based on the input task embedding $e$, ensuring the model chooses the appropriate high-instruction or low-action expert for the current task level. The  forward propagation process can be expressed as follows:
\vspace{-1em}
\begin{equation}
\ f(x) = W_0x + \Delta W x,
\end{equation}

\vspace{-3em}
\begin{equation}\label{eq:4}
\begin{split}
& \Delta W x= \sum_{i=1}^{N_1} G_1(x)_i \cdot E_i(x) \\
& + \sum_{i=1}^{N_2} G_2(e)_i \cdot E_i^h(x),
\end{split}
\end{equation}
where $ W_0 $ represents the pretrained weights of the base model, $ \Delta W $ indicates the weight updates produced by the LoRA-enhanced experts, $ N_1 $ denotes the number of token-level LoRA instances, and $ N_2 $ represents the number of task-level LoRA instances.Therefore, the total number of LoRA experts per layer $N$ can be expressed as $N = N_1 + N_2$. The token-level expert $ E(\cdot) $ and the task-level expert $ E^h(\cdot) $ compute their outputs using the following LoRA update rule:
\vspace{-0.5em}
\begin{equation}
\begin{cases}
E_i(x) = B_iA_ix_i ,\\
E_i^h(x) = B_i^hA_i^hx_i,
\end{cases}
\end{equation}
where $ B_i \in \mathbb{R}^{d \times r} $, $ A_i \in \mathbb{R}^{r \times k} $,  $B_i^h \in \mathbb{R}^{d \times r} $, $ A_i^h \in \mathbb{R}^{r \times k} $, the dimensions $ d $ and $ k $ correspond to the size of the pretrained weight matrix $ W_0 \in \mathbb{R}^{d \times k} $ in large language models. The parameter $ r $ denotes the low-rank dimension, with $r \ll \min(d, k) $. The router logits $ G_1(\cdot) $ denote the routing probabilities for each token-level expert, while $ G_2(\cdot) $ represent the routing probabilities for each task-level expert, which can be expressed as follows:
\begin{equation}\label{eq:6}
\begin{cases}
G_1(x) = \{i \mid p_i(x) \geqslant p_{(K)}(x)\} ,\\
G_2(e) = \{i \mid p_i(e) \geqslant p_{(1)}(e)\},
\end{cases}
\end{equation}
where $ p_i(\cdot) $ denotes the probability assigned to expert $ i $, $ p_{(K)}(\cdot) $ represents the $ K $-th highest probability, and $ p_{(1)}(\cdot) $ is the highest probability.






\begin{table*}[ht]
\belowrulesep=0pt
\aboverulesep=0pt
\centering
\caption{Low-action Incremental Comparison. The best value per metric is in
\textbf{bold}.}
\footnotesize 
\setlength{\tabcolsep}{2pt} 
\begin{tabularx}{\textwidth}{l*{8}{>{\centering\arraybackslash}X}|*{8}{>{\centering\arraybackslash}X}}
\toprule
\multicolumn{1}{c}{} & \multicolumn{8}{c|}{\textbf{LB}} & \multicolumn{8}{c}{\textbf{LE}} \\
\cmidrule(l){2-17}
\multirow{1}{*}{\textbf{Methods}} & \multicolumn{2}{c}{\textbf{Order1}} & \multicolumn{2}{c}{\textbf{Order2}} & \multicolumn{2}{c}{\textbf{Order3}}& \multicolumn{2}{c|}{\textbf{Avg}} & \multicolumn{2}{c}{\textbf{Order1}} & \multicolumn{2}{c}{\textbf{Order2}} & \multicolumn{2}{c}{\textbf{Order3}}& \multicolumn{2}{c}{\textbf{Avg}}  \\
& \textbf{AA$\uparrow$} & \textbf{FM$\downarrow$} & \textbf{AA$\uparrow$} & \textbf{FM$\downarrow$} & \textbf{AA$\uparrow$} & \textbf{FM$\downarrow$} & \textbf{AA$\uparrow$} & \textbf{FM$\downarrow$} & \textbf{AA$\uparrow$} & \textbf{FM$\downarrow$} & 
\textbf{AA$\uparrow$} & \textbf{FM$\downarrow$} & 
\textbf{AA$\uparrow$} & \textbf{FM$\downarrow$} & 
\textbf{AA$\uparrow$} & \textbf{FM$\downarrow$}\\
\midrule
EWC & 62.87 & 12.13 & 62.43 & 11.24 & 62.03 & 11.09 &62.44 &11.49 & 60.82 & 10.19 & 60.84 & 10.06 & 60.57 & 10.42&60.74&10.22  \\
CAMA & 62.89 & 11.53 & 62.96 & 11.50 & 62.71 & 12.02&62.85&11.68 & 61.41 & 10.14 & 60.30 & 10.64 & 60.36 & 10.81&60.69&10.53 \\
MoELoRA & 63.33 & 10.01 & 63.04 & 10.31 & 63.67 & 10.42 &63.35&10.25& 61.67 & 10.53 & 60.65 & 10.53 & 60.78 & 10.52&61.03&10.53 \\
O-LoRA & 64.61 & 8.85 & 64.53 & 7.70 & 64.68 & 8.62&64.61&8.39 & 63.06 & 7.91 & 62.20 & 8.05 & 62.75 & 8.14&62.67&8.03 \\
InfLoRA & 65.94 & 7.69 & 65.16 & 7.43 & 65.74 & 7.19&65.61&7.44 & 63.24 &7.04 & 62.32 & 7.90 & 62.24 & 7.97&62.60&7.64  \\
\textbf{Ours} & \textbf{67.75} & \textbf{3.83} & \textbf{67.52} & \textbf{3.09} & \textbf{68.45} & \textbf{3.18}&\textbf{67.91}&\textbf{3.37} & \textbf{65.43} & \textbf{3.61} & \textbf{64.42} & \textbf{4.05} & \textbf{64.78} & \textbf{3.71}&\textbf{64.88}&\textbf{3.79} \\
\bottomrule
\end{tabularx}
\label{tab:performance_comparison1}
\end{table*}

\begin{table*}[ht]
\belowrulesep=0pt
\aboverulesep=0pt
\centering
\caption{High instruction Incremental Comparison. The best value per metric is in \textbf{bold}.}
\footnotesize 
\setlength{\tabcolsep}{2pt} 
\begin{tabularx}{\textwidth}{l*{8}{>{\centering\arraybackslash}X}|*{8}{>{\centering\arraybackslash}X}}
\toprule
\multicolumn{1}{c}{} & \multicolumn{8}{c|}{\textbf{HB}} & \multicolumn{8}{c}{\textbf{HE}} \\
\cmidrule(l){2-17}
\multirow{1}{*}{\textbf{Methods}} & \multicolumn{2}{c}{\textbf{Order1}} & \multicolumn{2}{c}{\textbf{Order2}} & \multicolumn{2}{c}{\textbf{Order3}} & \multicolumn{2}{c|}{\textbf{Avg}}& \multicolumn{2}{c}{\textbf{Order1}} & \multicolumn{2}{c}{\textbf{Order2}} & \multicolumn{2}{c}{\textbf{Order3}}& \multicolumn{2}{c}{\textbf{Avg}}  \\
& \textbf{AA$\uparrow$} & \textbf{FM$\downarrow$} & \textbf{AA$\uparrow$} & \textbf{FM$\downarrow$} & \textbf{AA$\uparrow$} & \textbf{FM$\downarrow$} & \textbf{AA$\uparrow$} & \textbf{FM$\downarrow$} & \textbf{AA$\uparrow$} & \textbf{FM$\downarrow$} & \textbf{AA$\uparrow$} & \textbf{FM$\downarrow$}& \textbf{AA$\uparrow$} & \textbf{FM$\downarrow$}& \textbf{AA$\uparrow$} & \textbf{FM$\downarrow$} \\
\midrule
EWC & 52.03 & 10.39 & 51.24 & 10.70 & 51.37 & 11.39 &51.55&10.83& 53.59 & 11.30 & 53.60 & 10.78 & 53.03 & 11.87&53.41&11.32  \\
CAMA & 51.45 & 10.84 & 51.43 & 10.69 & 51.08 & 11.53&51.32&11.02 & 53.25 & 10.96 & 53.04 & 10.72 & 53.19 & 10.51&53.16&10.73 \\
MoELoRA & 51.99 & 10.95 & 52.29 & 10.31 & 51.96 & 10.59&52.08&10.62 & 53.84 & 12.88 & 53.87 & 11.20 & 54.00 & 11.02&53.90&11.70 \\
O-LoRA & 54.57 & 4.96 & 54.22 & 5.69 & 54.28 & 5.47&54.36&5.37 & 56.48 & 3.68 & 56.41 & 4.09 & 56.82 & 4.89&56.57&4.22 \\
InfLoRA & 54.52 & 4.60 & 54.82 & 4.12 & 54.17 & 5.03&54.50&4.58 & 56.56 & 3.30 & 56.88 & 4.29 & 56.49 & 4.53&56.64&4.04  \\
\textbf{Ours} & \textbf{55.90} & \textbf{2.89} & \textbf{55.34} & \textbf{3.40} & \textbf{55.51} & \textbf{3.07} & \textbf{55.58} & \textbf{3.12} & \textbf{58.01} & \textbf{1.62} & \textbf{57.84} & \textbf{1.97} & \textbf{57.21} & \textbf{1.69} & \textbf{57.69} & \textbf{1.76} \\
\bottomrule
\end{tabularx}
\label{tab:performance_comparison2}
\end{table*}

\begin{table*}[ht]
\belowrulesep=0pt
\aboverulesep=0pt
\centering
\caption{Hybrid Comparison. The best value per metric is in \textbf{bold}.}
\small
\begin{tabularx}{\textwidth}{l*{6}{>{\centering\arraybackslash}X}*{7}{>{\centering\arraybackslash}X}}
\toprule
\multirow{1}{*}{\textbf{Methods}} & \multicolumn{2}{c}{\textbf{Order1}} & \multicolumn{2}{c}{\textbf{Order2}} & \multicolumn{2}{c}{\textbf{Order3}}& \multicolumn{2}{c}{\textbf{Avg}} \\
& \textbf{AA$\uparrow$} & \textbf{FM$\downarrow$} & \textbf{AA$\uparrow$} & \textbf{FM$\downarrow$} & \textbf{AA$\uparrow$} & \textbf{FM$\downarrow$}& \textbf{AA$\uparrow$} & \textbf{FM$\downarrow$} \\
\midrule
EWC & 39.74 & 44.84 & 39.93 & 44.55 & 39.46 & 44.44&39.71&44.61 \\
CAMA & 39.59 & 43.51 & 39.88 & 43.90 & 39.76 & 43.55&39.74&43.65  \\
MoELoRA & 40.04 & 43.83 & 39.75 & 44.33 & 39.86 & 43.41&39.88&43.86 \\
O-LoRA & 43.89 & 37.99 & 44.70 & 38.42 & 43.15 & 37.53&43.91&37.98   \\
InfLoRA & 44.77 & 36.00 & 45.24 & 36.50 & 44.74 & 35.61&44.92&36.04   \\
\textbf{Ours} & \textbf{52.84} & \textbf{13.41} & \textbf{54.15} & \textbf{13.62} & \textbf{53.76} & \textbf{13.50} & \textbf{53.58} & \textbf{13.51} \\
\bottomrule
\end{tabularx}
\label{tab:performance_comparison3}
\end{table*}

\begin{algorithm}[!ht]
    \caption{Algorithm of Task-aware MoILE}
    \label{alg:AOS}
    \renewcommand{\algorithmicrequire}{\textbf{Input:}}
    \renewcommand{\algorithmicensure}{\textbf{Output:}}
    
    \begin{algorithmic}[1]
        \REQUIRE Pre-trained VLM parameters, number of
 tasks $T$, task data $\{D_t\}_{t=1}^T$.
        \ENSURE Updated model parameters.    
        \FOR{each task $t \in 1:T$}
            \STATE Obtain $W_p$ and $\overline{W}$ (Eqs.~(8),(9)).
            \FOR{each batch $D_{batch} \in D_t$}
                \STATE Update cluster centers $c_{\text{new}_j}$ (Eq.~(1)).
                \STATE Compute task embeddings $e$ (Eq.~(2)).
                \STATE Perform MoE forward pass (Eqs.~(3),(4).
                \STATE Calculate losses $L_s$, $L_o$ (Eqs.~(12), (14)).
                \STATE Update parameters $\overline{W}$.
            \ENDFOR
        \ENDFOR
        
        
        \RETURN Updated model parameters.
    \end{algorithmic}
\end{algorithm}

\subsection{Incremental LoRA}

To effectively prevent catastrophic forgetting and promote knowledge transfer across different tasks, we propose the Incremental LoRA method. Specifically, when updating LoRA expert parameters, we first perform Singular Value Decomposition (SVD) on the LoRA weight matrix $W$, which can be expressed as follows:
\vspace{-0.3em}
\begin{equation}
\ W = U \Sigma V^\top,
\end{equation}
where $\Sigma$ is a diagonal matrix whose entries along the diagonal are the singular values, $U$ and $V$ are orthogonal matrices consisting of the left and right singular vectors.

 We analyze the singular values after SVD and find that only a small portion of the LoRA parameters have significant values, while most have relatively low values. This indicates that a few parameters retain the core knowledge of the task. Based on this, we divide the LoRA expert parameters into principal component parameters and residual component parameters. The parameters corresponding to the top $p$ singular values are retained as the principal components $W_p$, while the remaining parameters are considered residual components $\overline{W}$:
\vspace{-1em}
\begin{equation}
\ W_p = U_p \Sigma_p V_p^\top,
\end{equation}

\vspace{-1em}
\begin{equation}
\overline{W} = W - W_p,
\end{equation}
where we freeze the principal component parameters $W_p$ and update only the residual parameters $\overline{W}$ during training. After merging the two, we obtain the LoRA parameters $\dot{W}$:
\vspace{-0.5em}
\begin{equation}
\ \dot{W} = \overline{W} + {W_p} ,
\end{equation}
\begin{equation}
\ W_p =B_pA_p, \overline{W} = \overline{B}\overline{A}.
\end{equation}

To prevent interference from new tasks from causing catastrophic forgetting, we introduce a LoRA singular value loss function $L_{s}$ during training. The singular value loss function aims to minimize the difference between the top $p$ singular values of the LoRA parameters $\dot{W}$ and those of the frozen principal component LoRA parameters $W_p$. Specifically, this loss function constrains the top $p$ singular values of the LoRA parameters $\dot{W}$ to be as close as possible to those of the frozen principal component LoRA parameters $W_p$. This ensures that the model retains knowledge from previous tasks, and the formulation is as follows:
\begin{equation}
\ L_{s} = \left| \sqrt{\sum_{i=1}^{p} \dot{\sigma_i}^2} - \sqrt{\sum_{i=1}^{p} \sigma_i^2} \right|,
\end{equation}
where $\dot{\sigma}$ denotes the singular values of the LoRA parameters $\dot{W}$, and $\sigma$ denotes the singular values of the frozen principal component LoRA parameters $W_p$. Additionally, we introduce an orthogonal loss $L_{\text{o}}$. Specifically, we apply an orthogonal constraint between the residual LoRA parameters $\overline{W}$ and the principal component LoRA parameters $W_p$ to reduce interference between them. The corresponding formula is as follows:
\vspace{-0.5em}
\begin{equation}
\ O = \overline{A}^TA_p ,
\end{equation}

\vspace{-1.5em}
\begin{equation}
\ L_{\text{o}} = \sum_{u,v} \|O[u,v]\|^2,
\end{equation}
where $O[u,v]$ denotes the element at the $u$-th row and $v$-th column of the matrix $O$. In summary, our total loss function $L_{\text{total}}$ is defined as follows:
\begin{equation}\label{eq:15}
L_{total} = L + \lambda_1 L_{s} + \lambda_2 L_{o},
\end{equation}
where $L$ is the next-token prediction loss of the large language model, $\lambda_1$ and $\lambda_2$ are hyperparameters that balance the contributions of each loss term. Our pseudocode is shown in \textbf{Algorithm \ref{alg:AOS}}.

\section{Experiments}
\subsection{Experimental Setting}

\textbf{Metrics.} We use Average Accuracy (AA) to evaluate the model's performance on tasks: 
\begin{equation}
    AA = \frac{1}{T} \sum_{t=1}^T A_{T,t},
\end{equation}
where $A_{T,t}$ is the performance on $t$-th task after training the task $T$. And we employ Forgetting Measure (FM) to measure the extent of catastrophic forgetting:
\vspace{-0.5em}
\begin{equation}
   f_{t,T} = \max_{i \in \{1,...,T-1\}}(A_{i,t} - A_{T,t}),\forall t < T,
\end{equation}
\vspace{-1em}
\begin{equation}
   FM_{T} = \frac{1}{T-1} \sum_{t=1}^{T-1} f_{t,T}.
\end{equation}

\textbf{Baseline.} We compare our method with state-of-the-art techniques in the continual learning domain, including EWC \cite{kirkpatrick2017overcoming}, CAMA \cite{kim2024online}, MoELoRA \cite{DBLP:conf/nips/ChenZLSSG24}, O-LoRA \cite{DBLP:conf/emnlp/WangCGXBZZGH23}, and InfLoRA \cite{liang2024inflora}. 

\textbf{Implementation details.} In our experiments, we employ the \textbf{AI2-THOR} environment alongside the ALFRED dataset \cite{shridhar2020alfred}. For each HEC setup, we train the agent using multiple randomly ordered sequences and evaluate it in unseen environments. For more details on these sequences, please refer to \textbf{Appendix \ref{sec:appendix_setup}}. The backbone language model is LLAMA 2-7B, and the vision encoder is set to CLIP ViT-L/14-336 by default. For additional details on the model training parameters, please refer to \textbf{Appendix \ref{sec:appendix_training}}.

\subsection{Comparison with State-of-the-art Methods}
\textbf{Table \ref{tab:performance_comparison1}} compares our method with other state-of-the-art approaches in the Low-level Action Incremental Learning setup. The results show that, in the LB and LE setups, our method outperforms the second-best by 2.3\% and 2.21\% in average accuracy, while reducing the forgetting measure by 4.07\% and 3.85\%. \textbf{Table \ref{tab:performance_comparison2}} presents the comparison in the High-level Instruction Incremental Learning setup. In the HB and HE setups, our method improves average accuracy by 1.08\% and 1.05\%, and reduces the forgetting measure by 1.46\% and 2.28\%. \textbf{Table \ref{tab:performance_comparison3}} shows the comparison in the Hybrid Hierarchical Incremental Learning setup. In this more practical setup, which closely aligns with real-world embodied intelligence tasks, our method boosts average accuracy by 8.66\% and reduces the forgetting measure by 22.53\%.
\vspace{-0.5em}
\begin{figure}[ht]
    \centering
    \includegraphics[width=0.48\textwidth]{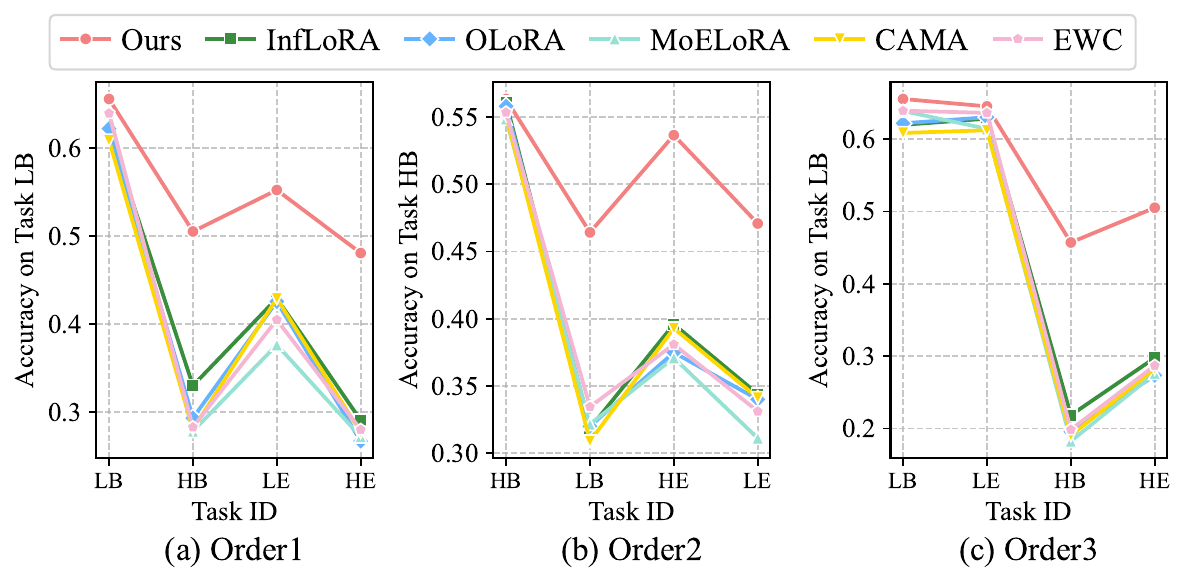}
    \caption{The performance trajectory of the first task in the continual learning process under different orders of the Hybrid Hierarchical Incremental Setup.}
    \label{fig:4}
\end{figure}

As shown in \textbf{Figure \ref{fig:4}}, with continuous learning, especially when learning different levels of knowledge (e.g., low-level actions and high-level instructions), our method effectively reduces the forgetting of previously learned knowledge. In contrast, the accuracy of other methods significantly declines. Compared to O-LoRA and InfLoRA, our method makes more accurate expert selections and better aligns with the tasks, thereby improving accuracy. When compared to MoELoRA, our method preserves existing knowledge while learning new knowledge, and, without knowing the task ID, it can choose the most suitable experts via clustering and task-level routing, thus improving accuracy and reducing forgetting.

\begin{table}[ht]
\belowrulesep=0pt
\aboverulesep=0pt
\centering
\caption{The Ablation Study of Different Components on the Hybrid Hierarchical Incremental Learning setup. \textbf{TR} stands for Task-level Router, \textbf{IL} represents Incremental LoRA, and \textbf{LS} and \textbf{LO} refer to Singular Value Loss and Orthogonal Loss.}
\small
\begin{tabularx}{0.48\textwidth}{l*{1}{>{\centering\arraybackslash}X}*{2}{>{\centering\arraybackslash}X}}
\toprule
\textbf{Methods}& \textbf{AA$\uparrow$} & \textbf{FM$\downarrow$} \\

\midrule
Ours - w/oIL & 45.53 & 38.68  \\
Ours - w/oTR & 47.87 & 19.45  \\
Ours - w/oLS & 49.10 & 21.05  \\
Ours - w/oLO & 49.31 & 22.72   \\
\textbf{Ours} & 53.58 & 13.51  \\
\bottomrule
\end{tabularx}
\label{tab:performance_comparison4}
\end{table}

\subsection{Ablation Study}

\vspace{-0.4em}
\begin{figure}[ht]
    \centering
    \includegraphics[width=0.48\textwidth]{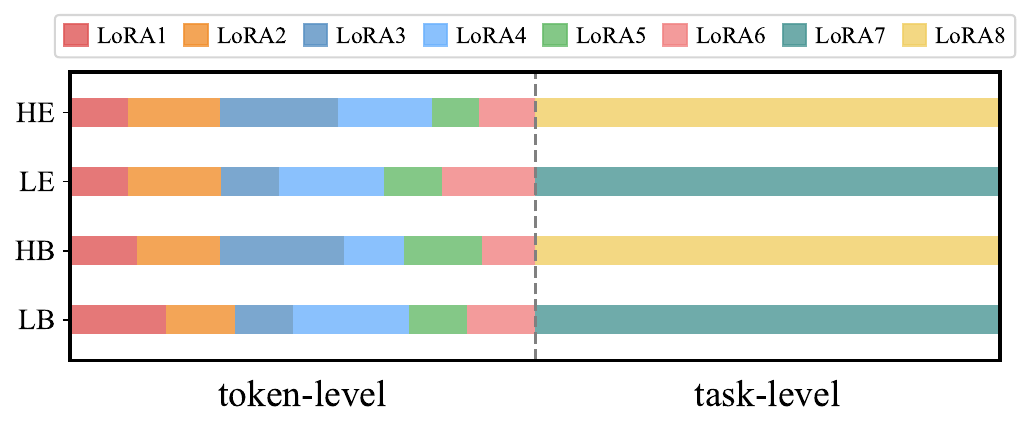}
    \caption{\textbf{Routing behavior across the token-level router and task-level router for different datasets.} The length of the color bar represents the router's preference for LoRA experts.}
    \label{fig:5}
\end{figure}


\textbf{Analysis of the individual components.} As shown in \textbf{Table \ref{tab:performance_comparison4}}, we conducted an ablation study on the Hybrid Hierarchical Incremental Learning setups to examine the role of each component in our method. "w/oIL", "w/oTR", "w/oLS", and "w/oLO" represent versions of the method that do not use Incremental LoRA, task-level router, singular value loss, and orthogonal loss, respectively. Compared to the full method, "w/oIL" shows a significant increase in task forgetting measure, indicating that Incremental LoRA plays a crucial role in preventing catastrophic forgetting. "w/oTR" exhibits a noticeable drop in average accuracy, highlighting the effectiveness of the task router in improving task accuracy. Both "w/oLS" and "w/oLO" show an increase in forgetting measure, suggesting $L_s$ and $L_o$ are important for mitigating catastrophic forgetting. These results validate the effectiveness of the individual components in our method.

\textbf{Analysis of Routing Results.} 
To further explore how our method alleviates the problem of catastrophic forgetting, we randomly selected a fixed number of samples from each dataset and visualized their LoRA block routing behaviors, as shown in \textbf{Figure \ref{fig:5}}. This reflects the choices of LoRA experts made by the token-level router and task-level router for different tasks, where the length of the color bar represents the router’s preference for each LoRA expert. LoRA1-6 represent token-level LoRA, and LoRA7-8 represent task-level LoRA. From the visualization, we can see that the token-level router makes relatively balanced choices for token-level LoRA, although there are some preferences for different tasks. On the other hand, the task-level router effectively selects the corresponding high-level instruction or low-level action LoRA experts based on the task hierarchy. For a comparison of different hyperparameters of our method, please refer to \textbf{Appendix \ref{sec:appendix_training}}.


\section{Related Work}

\textbf{Continual Learning.} Traditional continual learning methods can be categorized into memory-based \cite{rebuffi2017icarl,yan2021dynamically}, regularization-based \cite{DBLP:conf/naacl/WangLSL0LY24}, and dynamic expansion approaches \cite{xu2021adaptive}. Recently, parameter-efficient fine-tuning (PEFT) and mixture-of-experts (MoE) approaches have outperformed traditional methods in continual learning. PEFT methods \cite{yan2024effort,DBLP:conf/nips/QiaoM24} introduce a few parameters to minimize modifications, preserving past knowledge with limited flexibility. MoE-based methods \cite{li2024theory,dou2023loramoe,DBLP:journals/corr/abs-2312-12379} use router mechanisms for dynamic expert selection, achieving satisfactory results but still facing catastrophic forgetting. Our approach introduces a cluster-based hierarchical MoE structure for precise expert allocation without task-id supervision, while promoting knowledge transfer. Additionally, we propose Incremental LoRA, which retains the principal components of existing LoRA parameters and progressively learns new tasks, effectively addressing continual learning challenges.


\textbf{Lifelong Learning for Embodied Agents.} Embodied continual learning enables agents to continuously learn new tasks and adapt their behavior in dynamic environments \cite{DBLP:journals/inffus/LesortLSMFR20}, while preserving previously acquired knowledge. While existing research on manipulation tasks primarily focuses on reinforcement learning \cite{DBLP:conf/nips/WolczykZPKM21} and imitation learning \cite{DBLP:conf/nips/MendezSE18,DBLP:conf/iros/GaoGGZ021}, vision-language navigation (VLN) has explored incremental learning approaches. Kim et al. \cite{kim2024online} introduced Behavior Incremental Learning and Environment Incremental Learning for VLN, aiming to equip agents with new skills and adaptability to new environments. However, these methods mainly focus on low-level action learning, neglecting the importance of high-level planning and decision-making. To address this, we introduce Hierarchical Embodied Continual Learning Setups, proposing High-level Instruction Incremental Learning and Low-level Action Incremental Learning for VLN, enabling agents to effectively learn high-level instructions for task planning and decision-making, while retaining low-level action execution capabilities, advancing embodied intelligence continual learning.

\section{Conclusion}
This paper addresses the challenges of continual learning in embodied agents driven by LLMs, and proposes an innovative embodied continual learning setups and method. The main contributions include: First, we design the Hierarchical Embodied Continual Learning Setups that decomposes the learning process into high-level instructions and low-level actions, and define five incremental learning setups to enhance the agent's ability for autonomous continual learning. Second, the proposed continual learning method, Task-aware MoILE, combines cross-modal task clustering, a hierarchical task routing mechanism, and incremental LoRA, enabling the agent to efficiently select task-specific experts without prior knowledge of task IDs, while ensuring compatibility and minimal interference between new and old tasks. Through detailed comparative experiments and ablation studies, we validate the effectiveness of each component in preventing catastrophic forgetting and improving task execution accuracy. 
\section{Limitations}
Despite our study making significant strides by introducing Hierarchical Embodied Continual Learning Setups and the Task-aware MoILE method to address catastrophic forgetting in cross-level learning for agents, there are several key areas that warrant further exploration and improvement in future work: Firstly, our research has only been experimentally validated on the ALFRED dataset. To more comprehensively assess the effectiveness and generalizability of our approach, future studies should aim to extend its application to more complex and varied real-world scenarios. Secondly, the current experiments rely on existing high-quality datasets for training and testing. However, in the practical application scenarios of embodied intelligence, it is often challenging to obtain large amounts or high-quality data. Therefore, an important direction for future research is to explore how to apply our method in few-shot or semi-supervised continual learning environments. This would not only broaden the applicability of the method but also enhance the adaptability and learning efficiency of agents in data-scarce conditions.
\section{Acknowledgements}
This work was supported by the National Key Research and Development Program of China (Youth Scientist Project) under Grant No. 2024YFB4504300 and the Shenzhen-Hong Kong Joint Funding Project (Category A) under Grant No. SGDX20240115103359001.

\bibliography{custom}

\appendix

\section{Details of Hierarchical Embodied Continual Learning Setups}
\label{sec:appendix_setup}

\subsection{High-level Instruction Behavior Incremental Learning}
\label{sec:behavior_incremental_learning}

In our setups, there are seven distinct behavior categories: EXAMINE, PICK\&PLACE, HEAT, COOL, CLEAN, PICK2\&PLACE, and MOVABLE. In the high-level Instruction Behavior Incremental Learning setup, the model is required to learn high-level planning under different behavior types. The total dataset for this setup comprises 187,103 instances. We randomly ordered these behavior categories to generate the following three sequences for model training.

\begin{enumerate}
    \item EXAMINE $\rightarrow$ HEAT $\rightarrow$ PICK2\&PLACE $\rightarrow$ COOL $\rightarrow$ PICK\&PLACE $\rightarrow$ CLEAN $\rightarrow$ MOVABLE
    \item PICK2\&PLACE $\rightarrow$ CLEAN $\rightarrow$ MOVABLE $\rightarrow$ PICK\&PLACE $\rightarrow$ HEAT $\rightarrow$ EXAMINE $\rightarrow$ COOL
    \item MOVABLE $\rightarrow$ COOL $\rightarrow$ PICK\&PLACE $\rightarrow$ HEAT $\rightarrow$ CLEAN $\rightarrow$ EXAMINE $\rightarrow$ PICK2\&PLACE
\end{enumerate}

\subsection{High-level Instruction Environment Incremental Learning}
\label{sec:environment_incremental_learning}

In the dataset, four unique environments are included: KITCHENS, LIVINGROOMS, BEDROOMS, and BATHROOMS. In the high-level Instruction Environment Incremental Learning setup, the model is tasked with acquiring the ability to perform high-level planning across varying environmental contexts. The total dataset for for this setup comprises 81048 instances. We also randomly ordered these environments to generate the following three sequences for model training.

\begin{enumerate}
    \item BEDROOMS $\rightarrow$ BATHROOMS $\rightarrow$ LIVINGROOMS $\rightarrow$ KITCHENS
    \item KITCHENS $\rightarrow$ BEDROOMS $\rightarrow$ LIVINGROOMS $\rightarrow$ BATHROOMS
    \item BATHROOMS $\rightarrow$ LIVINGROOMS $\rightarrow$ KITCHENS $\rightarrow$ BEDROOMS
\end{enumerate}

\subsection{Low-level Action Behavior Incremental Learning}

In the Low-level Action Behavior Incremental Learning setup, the model is required to acquire low-level control capabilities across diverse behavior types. We trained the model using the same sequences as described above in \ref{sec:behavior_incremental_learning}.

\begin{enumerate}
    \item EXAMINE $\rightarrow$ HEAT $\rightarrow$ PICK2\&PLACE $\rightarrow$ COOL $\rightarrow$ PICK\&PLACE $\rightarrow$ CLEAN $\rightarrow$ MOVABLE
    \item PICK2\&PLACE $\rightarrow$ CLEAN $\rightarrow$ MOVABLE $\rightarrow$ PICK\&PLACE $\rightarrow$ HEAT $\rightarrow$ EXAMINE $\rightarrow$ COOL
    \item MOVABLE $\rightarrow$ COOL $\rightarrow$ PICK\&PLACE $\rightarrow$ HEAT $\rightarrow$ CLEAN $\rightarrow$ EXAMINE $\rightarrow$ PICK2\&PLACE
\end{enumerate}

\subsection{Low-level Action Environment Incremental Learning}

In the Low-level Action Environment Incremental Learning setup, the model is required to acquire low-level control capabilities across diverse environmental contexts. We trained the model using the same sequences as described above in \ref{sec:environment_incremental_learning}.

\begin{enumerate}
    \item BEDROOMS $\rightarrow$ BATHROOMS $\rightarrow$ LIVINGROOMS $\rightarrow$ KITCHENS
    \item KITCHENS $\rightarrow$ BEDROOMS $\rightarrow$ LIVINGROOMS $\rightarrow$ BATHROOMS
    \item BATHROOMS $\rightarrow$ LIVINGROOMS $\rightarrow$ KITCHENS $\rightarrow$ BEDROOMS
\end{enumerate}

\subsection{Hybrid Hierarchical Incremental Learning}

In our experimental setup, we also designed three distinct sequences comprising four incremental learning modes, resulting in the following orders for model training.

\begin{enumerate}
    \item Low-level Action Behavior $\rightarrow$ High-level Instruction Behavior $\rightarrow$ Low-level Action Environment $\rightarrow$ High-level Instruction Environment
    \item High-level Instruction Behavior $\rightarrow$ Low-level Action Behavior $\rightarrow$ High-level Instruction Environment $\rightarrow$ Low-level Action Environment
    \item Low-level Action Behavior $\rightarrow$ Low-level Action Environment $\rightarrow$ High-level Instruction Behavior $\rightarrow$ High-level Instruction Environment
\end{enumerate}

\section{Details of Model Training}
\label{sec:appendix_training}
In this section, we provide the detailed hyperparameter settings used for training the proposed model. The hyperparameters are set as follows. We conducted the training using two NVIDIA RTX 4090 D GPUs. A complete training cycle of the Hybrid Hierarchical Incremental Learning setup typically requires approximately 8 hours.

\begin{itemize}
    \item In Equation \ref{eq:2}, $M$ is set to 4, where $M$ denotes the number of clusters.
    \item In Equation \ref{eq:4}, $N_1$ is set to 6, and $N_2$ is set to 2, where $N_1$ represents the number of token-level LoRA experts, and $N_2$ represents the number of task-level LoRA experts.
    \item In Equation \ref{eq:6}, $K$ is set to 2, where $K$ denotes the number of selected token-level LoRA experts in the top-$K$ gating mechanism of the MoE layer.
    \item In Equation \ref{eq:15}, $\lambda_1$ is set to 1.0, and $\lambda_2$ is set to 0.5, where $\lambda_1$ is the weight for the LoRA singular value loss, and $\lambda_2$ is the weight for the orthogonal loss.
\end{itemize}

Additionally, we conducted experiments with different hyperparameter settings to evaluate the model's performance.

\begin{table}[ht]
\belowrulesep=0pt
\aboverulesep=0pt
\centering
\caption{Model performance (AA and FM) under different hyperparameter settings.}
\small
\renewcommand{\arraystretch}{1.6} 
\begin{tabular}{cccccccc}
\toprule
\textbf{\(N_1\)} & \textbf{\(N_2\)} & \textbf{\(K\)} & \textbf{\(\lambda_1\)} & \textbf{\(\lambda_2\)} & \textbf{\(M\)} & \textbf{AA}$\uparrow$ & \textbf{FM}$\downarrow$ \\
\midrule
2 & 2 & 1 & 1.0 & 0.5 & 4 & 48.10 & 18.56 \\
6 & 2 & 2 & 0.5 & 1.0 & 4 & 53.10 & 14.27 \\
6 & 2 & 2 & 1.0 & 0.5 & 8 & 52.30 & 15.84 \\
6 & 2 & 2 & 1.0 & 0.5 & 4 & \textbf{53.58} & \textbf{13.51}  \\
\bottomrule
\end{tabular}
\label{tab:hyperpara_test}
\end{table}

The results summarized in \textbf{Table \ref{tab:hyperpara_test}} indicate that the current hyperparameter setting achieves optimal performance. Reducing token-level experts to $N_1=2$ with $K=1$ degrades AA by 5.48\%, suggesting insufficient capacity for token-level adaptation. Setting the LoRA singular value loss weight ($\lambda_1=0.5$) smaller than the orthogonal loss weight ($\lambda_2=1.0$) results in slightly degraded performance. This suggests that a stronger singular value constraint is more effective in balancing parameter diversity and adaptation stability. Setting cluster number $M$ to 8 increases FM by 2.33\% and degrades AA by 1.28\%, implying over-clustering introduces noisy task-feature disentanglement. These results demonstrate that our final hyperparameter setting balances these factors.

\begin{table}[ht]
\belowrulesep=0pt
\aboverulesep=0pt
\centering
\small
\caption{Comprehensive analysis of computational overheads.}
\label{tab:computational_costs_comparison}
\begin{tabular}{lcc}
\hline
\textbf{Method}                &\textbf{Memory (GB)} & \textbf{Training Time (h)} \\
\hline
EWC                    & 15.37                 & 5.2               \\

CAMA                     & 15.38                 & 5.8               \\

MoELoRA                   & 16.32                 & 6.8               \\

O-LoRA                    & 19.26                 & 7.9               \\

InfLoRA                   & 19.58                 & 7.2               \\

Ours                      & 16.06                 & 7.6               \\
\hline
\end{tabular}
\end{table}

\begin{table}[ht]
\belowrulesep=0pt
\aboverulesep=0pt
\centering
\small
\caption{Performance comparison under different numbers of clusters.}
\label{tab:cluster_performance}
\begin{tabular}{ccc} 
\toprule
\textbf{Number of Clusters} & \textbf{AA$\uparrow$} & \textbf{FM$\downarrow$} \\
\midrule
2                  & 53.27        & 14.03          \\
4                  & 53.58        & 13.51          \\
8                  & 52.30        & 15.84          \\
\bottomrule
\end{tabular}
\end{table}

As demonstrated in \textbf{Table~\ref{tab:computational_costs_comparison}}, our proposed method achieves the highest task accuracy and the lowest knowledge forgetting (see Tables~1, 2, and 3 in the paper for detailed comparisons) while maintaining memory efficiency and inference efficiency comparable to the baseline method MoELoRA. Notably, this performance advantage is accompanied by a moderate increase in training time.

In designing the cross-modal clustering module, our primary objective was to enable the model to effectively distinguish between high-level instructions and low-level actions. To this end, we employed a fixed number of clusters. \textbf{Table~\ref{tab:cluster_performance}} illustrates the impact of varying cluster numbers on model performance under consistent experimental conditions. As the results show, the number of clusters exhibits minimal influence on overall performance. Based on these findings, we adopted a fixed number of clusters in practical applications. Specifically, setting the number of clusters to 2 is sufficient to differentiate between high-level instructions and low-level actions, ensuring both simplicity and effectiveness in real-world scenarios.

\section{Analysis of LoRA Parameters and Visual-Text Embeddings}
\label{sec:appendix_analysis}

\begin{figure}[ht]
     \centering
     \includegraphics[width=0.48\textwidth]{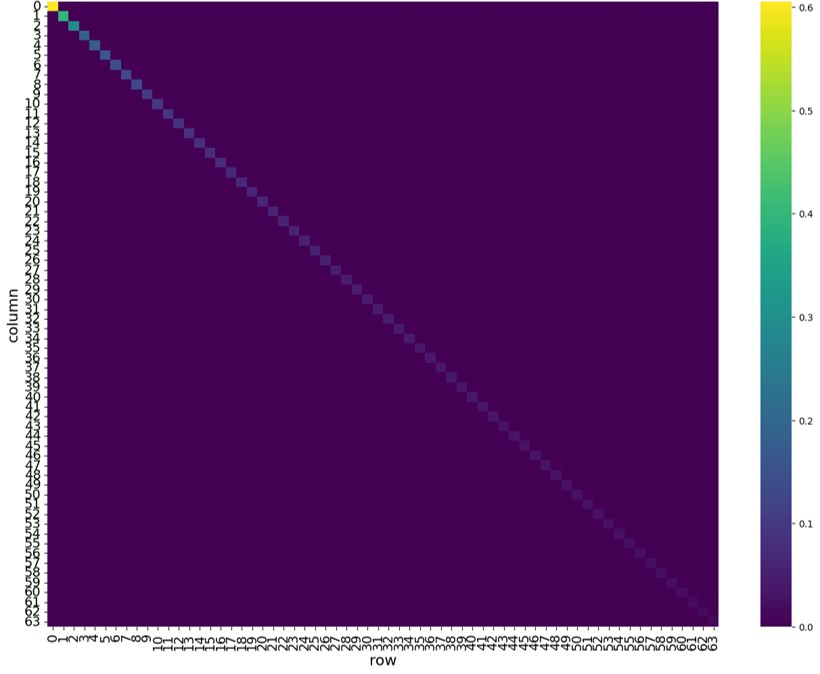}
     \caption{The singular values of the LoRA parameters after SVD.}
     \label{fig:6}
\end{figure}

As shown in \textbf{Figure \ref{fig:6}}, we analyze the singular values after SVD and find that only a small portion of the LoRA parameters have significant values, while most have relatively low values. This indicates that a few parameters retain the core knowledge of the task. Based on this, we divide the LoRA expert parameters into principal component parameters and residual component parameters to prioritize critical information during model updates.

\begin{figure}[ht]
     \centering
     \includegraphics[width=0.48\textwidth]{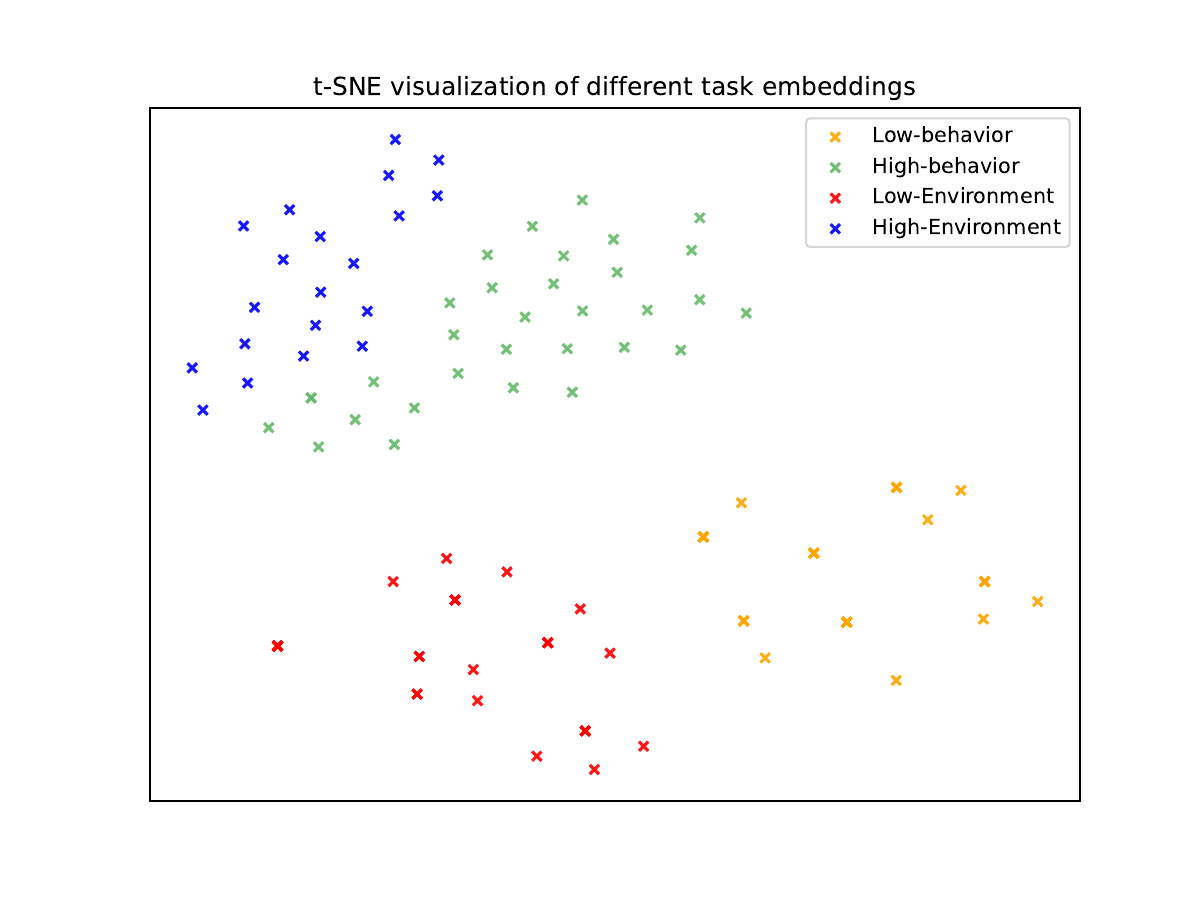}
     \caption{\textbf{T-SNE visualization of visual-text embeddings.} Different colors indicating different hierarchical continual learning datasets.}
     \label{fig:7}
\end{figure}

As shown in \textbf{Figure \ref{fig:7}}, we use t-SNE to visualize the visual-text embeddings of different datasets. It is evident that tasks of the same type are clustered together in the feature space, while tasks with similar hierarchical levels, such as High-behavior and High-Environment under High-level Instruction, are more closely distributed. This geometric arrangement provides insights for hierarchical knowledge preservation.

\end{document}